\documentclass[11pt,a4paper]{article}
\usepackage[hyperref]{acl2019}
\usepackage{times}
\usepackage{latexsym}
\usepackage{graphicx}
\usepackage{url}
\usepackage{amsmath}
\usepackage{breqn}
\usepackage{makecell}
\usepackage{tabularx}
\usepackage{bm}
\DeclareMathOperator*{\argmax}{argmax} %
\newcommand{\specialcell}[2][c]{%
\begin{tabular}[#1]{@{}c@{}}#2\end{tabular}}

\aclfinalcopy % Uncomment this line for the final submission
\title{Multi-News: a Large-Scale Multi-Document Summarization \\ Dataset and Abstractive Hierarchical Model}

\author{Alexander R. Fabbri \qquad Irene Li \vspace{0.005\textwidth}  \\  \qquad \textbf{Tianwei She} \qquad \textbf{Suyi Li}    \quad \hspace{0.01\textwidth} \textbf{Dragomir R. Radev}  \AND
  {\normalfont Department of Computer Science} \\
   {\normalfont Yale University} \\
{{ \normalfont \{alexander.fabbri,irene.li,tianwei.she,suyi.li,dragomir.radev}}\}@yale.edu \AND \vspace{-1.6cm}}

\date{}

\begin{document}
\maketitle
\begin{abstract}
Automatic generation of summaries from multiple news articles is a valuable tool as the number of online publications grows rapidly. Single document summarization (SDS) systems have benefited from advances in neural encoder-decoder model thanks to the availability of large datasets. However, multi-document summarization (MDS) of news articles has been limited to datasets of a couple of hundred examples. In this paper, we introduce Multi-News, the first large-scale MDS news dataset. Additionally, we propose an end-to-end model which incorporates a traditional extractive summarization model with a standard SDS model and achieves competitive results on MDS datasets. We benchmark several methods on Multi-News and release our data and code in hope that this work will promote advances in summarization in the multi-document setting\footnote{\url{https://github.com/Alex-Fabbri/Multi-News}}. 
\end{abstract}

%  introduce the paper idea, dataset + end-to-end model + benchmarks
\section{Introduction}
%  introduce data sparsity problem, recent attempts to deal with it, how we deal with it with our dataset, approaches to MDS and how we innovate, and the fact that we benchmark this new dataset
%  three contributions:
% dataset
% MMR method
% Summarization raises a general interest as there are various applications like news headline generation. Downstream tasks like survey generation and resum\'e generation also benefit from the summarization models.  However, neural abstractive summarization has been focused on SDS tasks \cite{Rush:15, see2017ptr_gen, Gehrmann:18}.  

Summarization is a central problem in Natural Language Processing with increasing applications as the desire to receive content in a concise and easily-understood format increases. Recent advances in neural methods for text summarization have largely been applied in the setting of single-document news summarization and headline generation \cite{Rush:15, see2017ptr_gen, Gehrmann:18}.
These works take advantage of large datasets such as the Gigaword Corpus \cite{napoles12giga}, the CNN/Daily Mail (CNNDM) dataset \cite{Hermann:15}, the New York Times dataset \cite{NYT} and the Newsroom corpus \cite{Grusky:18}, which contain on the order of hundreds of thousands to millions of article-summary pairs. However, multi-document summarization (MDS), which aims to output summaries from document clusters on the same topic, has largely been performed on datasets with less than 100 document clusters such as the DUC 2004 \cite{paul2004introduction} and TAC 2011 \cite{owczarzak2011} datasets, and has benefited less from advances in deep learning methods. 

\begin{table}[t!]
\centering
\small
\begin{tabularx}{\columnwidth}{|X|}
\hline
\textbf{Source 1} \\ \hline
\textcolor{red}{Meng Wanzhou, Huawei's chief financial officer and deputy chair, was arrested in Vancouver on 1 December.} Details of the arrest have not been released...\\

\hline

\textbf{Source 2} \\ \hline
\textcolor{blue}{A Chinese foreign ministry spokesman said on Thursday that Beijing had separately called on the US and Canada to \lq\lq clarify the reasons for the detention \rq\rq  immediately and \lq\lq immediately release the detained person \rq\rq.} The spokesman...\\

\hline
\textbf{Source 3}\\ \hline
\textcolor{red}{Canadian officials have arrested Meng Wanzhou}, the chief financial officer and deputy chair of the board for the Chinese tech giant Huawei,...Meng was arrested in Vancouver on Saturday and \textcolor{purple}{is being sought for extradition by the United States.} A bail hearing has been set for Friday...\\
\hline
\hline
\textbf{Summary} \\ \hline
...Canadian authorities say \textcolor{purple}{she was being sought for extradition to the US}, where the company is being investigated for possible violation of sanctions against Iran. Canada's justice department said \textcolor{red}{Meng was arrested in Vancouver on Dec. 1}... China's embassy in Ottawa released a statement….. \textcolor{blue}{\lq\lq The Chinese side has lodged stern representations with the US and Canadian side, and urged them to immediately correct the wrongdoing \rq\rq and restore Meng's freedom}, the statement said...\\
\hline
\end{tabularx}
\caption{An example from our multi-document summarization dataset showing the input documents and their summary. Content found in the summary is color-coded.}
\label{tab:example}
\end{table}

% summary: http://www.newser.com/story/268216/canada-arrests-cfo-of-china-tech-giant.html
% art1:https://www.bbc.com/news/business-46462858
% art2:https://www.theguardian.com/technology/2018/dec/05/meng-wanzhou-huawei-cfo-arrested-vancouver
% art3:https://www.cbc.ca/news/politics/canada-huawei-arrest-cfo-1.4934269

 Multi-document summarization of news events offers the challenge of outputting a well-organized summary which covers an event comprehensively while simultaneously avoiding redundancy.  The input documents may differ in focus and point of view for an event. We present an example of multiple input news documents and their summary in  Figure \ref{tab:example}. The three source documents discuss the same event and contain overlaps in content: the fact that \textit{Meng Wanzhou was arrested} is stated explicitly in Source 1 and 3 and indirectly in Source 2. However, some sources contain information not mentioned in the others which should be included in the summary: Source 3 states that \textit{(Wanzhou) is being sought for extradition by the US} while only Source 2 mentioned \textit{the attitude of the Chinese side}.

Recent work in tackling this problem with neural models has attempted to exploit the graph structure among discourse relations in text clusters \cite{yasunaga17graph} or through an auxiliary text classification task \cite{17classification}. Additionally, a couple of recent papers have attempted to adapt neural encoder decoder models trained on single document summarization datasets to MDS \cite{lebanoff18mds,baumel18mds, Zhang:18}.

However, data sparsity has largely been the bottleneck of the development of neural MDS systems.  The creation of large-scale multi-document summarization dataset for training has been restricted due to the sparsity and cost of human-written summaries. \newcite{liu18wikisum} trains abstractive sequence-to-sequence models on a large corpus of Wikipedia text with citations and search engine results as input documents. However, no analogous dataset exists in the news domain.  To bridge the gap, we introduce Multi-News, the first large-scale MDS news dataset, which contains 56,216 articles-summary pairs. We also propose a hierarchical model for neural abstractive multi-document summarization, which consists of a pointer-generator network \cite{see2017ptr_gen} and an additional Maximal Marginal Relevance (MMR) \cite{carbonell1998use} module that calculates sentence ranking scores based on relevancy and redundancy. We integrate sentence-level MMR scores into the pointer-generator model to adapt the attention weights on a word-level. Our model performs competitively on both our Multi-News dataset and the DUC 2004 dataset on ROUGE scores. We additionally perform human evaluation on several system outputs.

Our contributions are as follows: We introduce the first large-scale multi-document summarization datasets in the news domain. We propose an end-to-end method to incorporate MMR into pointer-generator networks. Finally, we benchmark various methods on our dataset to lay the foundations for future work on large-scale MDS.

\section{Related Work}
% Alex
% Here we will introduce MDS (non-neural methods -- extractive and abstractive), note success of neural methods in single document setting (all classic papers) and then go to the recent attempts to 
Traditional non-neural approaches to multi-document summarization have been both extractive \cite{carbonell1998use, radev00centroid, erkan2004lexrank, mihalcea2004textrank, haghighi09content} as well as abstractive \cite{McKeown&Radev95, radev98mds, barzilay99fusion, ganesan10opinosis}. 
%\par
Recently, neural methods have shown great promise in text summarization, although largely in the single-document setting, with both extractive \cite{nallapai16b, cheng16ext, narayan18rl} and abstractive methods \cite{chopra16og, nallapati16a, see2017ptr_gen, paulus17rl, cohan18discourse, celikyilmaz18rl, Gehrmann:18}

In addition to the multi-document methods described above which address data sparsity, recent work has attempted unsupervised and weakly supervised methods in non-news domains \cite{liu18unsupervised, angelidis18opinions}. The methods most related to this work are SDS adapted for MDS data. 
\newcite{zhang18mds} adopts a hierarchical encoding framework trained on SDS data to MDS data by adding an additional document-level encoding. \newcite{baumel18mds} incorporates query relevance into standard sequence-to-sequence models. \newcite{lebanoff18mds} adapts encoder-decoder models trained on single-document datasets to the MDS case by introducing an external MMR module which does not require training on the MDS dataset. In our work, we incorporate the MMR module directly into our model, learning weights for the similarity functions simultaneously with the rest of the model.
% and using distributed representations for sentence representations. 
% Our work, however, focuses on the news domain and the adaptation of pointer-generator networks with MMR in an end-to-end manner. Neural abstractive MDS has also recently been explored in both a weakly-supervised \cite{angelidis18opinions} and unsupervised \cite{liu18unsupervised} manner.
\section{Multi-News Dataset}
Our dataset, which we call Multi-News, consists of news articles and human-written summaries of these articles from the site newser.com. Each summary is professionally written by editors and includes links to the original articles cited. We will release stable Wayback-archived links, and scripts to reproduce the dataset from these links. Our dataset is notably the first large-scale dataset for MDS on news articles. Our dataset also comes from a diverse set of news sources; over 1,500 sites appear as source documents 5 times or greater, as opposed to previous news datasets (DUC comes from 2 sources, CNNDM comes from CNN and Daily Mail respectively, and even the Newsroom dataset \cite{Grusky:18} covers only 38 news sources). A total of 20 editors contribute to 85\% of the total summaries on newser.com. Thus we believe that this dataset allows for the summarization of diverse source documents and summaries.

% Introduce the dataset, the source -- need to do analysis on the number of sources from which we get the dataset
\subsection{Statistics and Analysis}
The number of collected Wayback links for summaries and their corresponding cited articles totals over 250,000. We only include examples with between 2 and 10 source documents per summary, as our goal is MDS, and the number of examples with more than 10 sources was minimal. The number of source articles per summary present, after downloading and processing the text to obtain the original article text, varies across the dataset, as shown in Table \ref{tab:num_sources}. We believe this setting reflects real-world situations; often for a new or specialized event there may be only a few news articles. Nonetheless, we would like to summarize these events in addition to others with greater news coverage. 

% \begin{table}[h!]
% \small
% \centering
% %\small
% \begin{tabular}{c c} \Xhline{2\arrayrulewidth}
%              \textbf{\# of source articles} & \textbf{Frequency} \\ \Xhline{\arrayrulewidth}
%              2 & 23,894  \\ %\Xhline{\arrayrulewidth}
%              3 & 12,707 \\ %\Xhline{\arrayrulewidth}
%              4 & 5,022 \\ %\Xhline{\arrayrulewidth}
%              5 & 1,873  \\ %\Xhline{\arrayrulewidth}
%              6 & 763  \\ %\Xhline{\arrayrulewidth}
%              7 & 382  \\ %\Xhline{\arrayrulewidth}
%              8 & 209  \\ %\Xhline{\arrayrulewidth}
%              9 & 89  \\ %\Xhline{\arrayrulewidth}
%              10 & 33  \\ \Xhline{2\arrayrulewidth}
% \end{tabular}
% \caption{The number of source articles per summary by frequency in our dataset.}
% \label{tab:num_sources}
% \end{table}

\begin{table}[t!]
\small
\centering
%\small
\begin{tabular}{c c | cc} \Xhline{2\arrayrulewidth}
             \textbf{\# of source} & \textbf{Frequency}& \textbf{\# of source} & \textbf{Frequency} \\ \Xhline{\arrayrulewidth} 
             
             2	&	23,894	&	7	&	382	\\
            3	&	12,707	&	8	&	209	\\
            4	&	5,022	&	9	&	89	\\
            5	&	1,873	&	10	&	33	\\
            6	&	763	&		&		\\
              
             \Xhline{2\arrayrulewidth}
            
\end{tabular}
\caption{The number of source articles per example, by frequency, in our dataset.}
\label{tab:num_sources}
\end{table}

% TODO concat or individual documents -- 
We split our dataset into training (80\%,
44,972), validation (10\%, 5,622), and test (10\%,
5,622) sets. Table \ref{tab:statistics} compares Multi-News to other news datasets used in experiments below. We choose to compare Multi-News with DUC data from 2003 and 2004 and TAC 2011 data, which are typically used in multi-document settings. Additionally, we compare to the single-document CNNDM dataset, as this has been recently used in work which adapts SDS to MDS \cite{lebanoff18mds}. The number of examples in our Multi-News dataset is two orders of magnitude larger than previous MDS news data. The total number of words in the concatenated inputs is shorter than other MDS datasets, as those consist of 10 input documents, but larger than SDS datasets, as expected. Our summaries are notably longer than in other works, about 260 words on average. While compressing information into a shorter text is the goal of summarization, our dataset tests the ability of abstractive models to generate fluent text concise in meaning while also coherent in the entirety of its generally longer output, which we consider an interesting challenge. 
% Furthermore, statistics in Table \ref{tab:statistics} emphasize the size of input compared, although the comparison is only fair between our dataset and DUC04, as the two consist of multiple documents. 
 \begin{table*}[t]
\centering
\small
\begin{tabular}{c c c c c c c} \Xhline{2\arrayrulewidth}
             \textbf{Dataset} & \textbf{\# pairs} & \specialcell{\textbf{\# words} \\\textbf{(doc)}} & \specialcell{\textbf{\# sents} \\\textbf{(docs)}} & \specialcell{\textbf{\# words} \\\textbf{(summary)}} &  \specialcell{\textbf{\# sents} \\\textbf{(summary)}} & \textbf{vocab size} \\ \Xhline{\arrayrulewidth}
             Multi-News & 44,972/5,622/5,622 & 2,103.49 & 82.73 & 263.66 & 9.97 &666,515 \\ %\Xhline{2\arrayrulewidth}
            DUC03+04 & 320 & 4,636.24 & 173.15 & 109.58 & 2.88 & 19,734 \\ %\Xhline{2\arrayrulewidth}
             TAC 2011 & 176 & 4,695.70 & 188.43  & 99.70 & 1.00 &24,672 \\ %\Xhline{2\arrayrulewidth}
            CNNDM & 287,227/13,368/11,490 & 810.57 & 39.78  & 56.20 & 3.68 &717,951 \\ \Xhline{2\arrayrulewidth}
\end{tabular}
\caption{Comparison of our Multi-News dataset to other MDS datasets as well as an SDS dataset used as training data for MDS (CNNDM). Training, validation and testing size splits (article(s) to summary) are provided when applicable. Statistics for multi-document inputs are calculated on the concatenation of all input sources.}
\label{tab:statistics}
\end{table*}

% \begin{figure}[t]
%     \centering
%     \includegraphics[width=\columnwidth]{plots/document_counts.png}
%     \caption{Number of source articles per summary}
%     \label{fig:mmr}
%     % \vspace{-2mm}
% \end{figure} 

\begin{table}[t!]
\centering
\small
\resizebox{\columnwidth}{!}{\begin{tabular}{c c c c c} \Xhline{2\arrayrulewidth}
             \specialcell{\textbf{\% novel} \\ \textbf{n-grams}} & \textbf{Multi-News} & \textbf{DUC03+04} & \textbf{TAC11} & \textbf{CNNDM} \\ \Xhline{\arrayrulewidth}
             uni-grams & 17.76 & 27.74 & 16.65 & 19.50  \\ %\Xhline{2\arrayrulewidth}
             bi-grams & 57.10 & 72.87 & 61.18  & 56.88 \\ %\Xhline{2\arrayrulewidth}
             tri-grams & 75.71 & 90.61 & 83.34  & 74.41 \\ %\Xhline{2\arrayrulewidth}
             4-grams & 82.30 & 96.18 & 92.04  & 82.83 \\ \Xhline{2\arrayrulewidth}
\end{tabular}}
\caption{Percentage of n-grams in summaries which do not appear in the input documents , a measure of the abstractiveness, in relevant datasets.}
\label{tab:ngrams}
\end{table}

%  \ref{tab:statistics}. 
% (Alex)
%  introduce stats about average words, summaries, sentences, how extractive it is, 
\subsection{Diversity}  \label{diversity_section}
\begin{figure}[t]
    \centering
    \includegraphics[width=\linewidth]{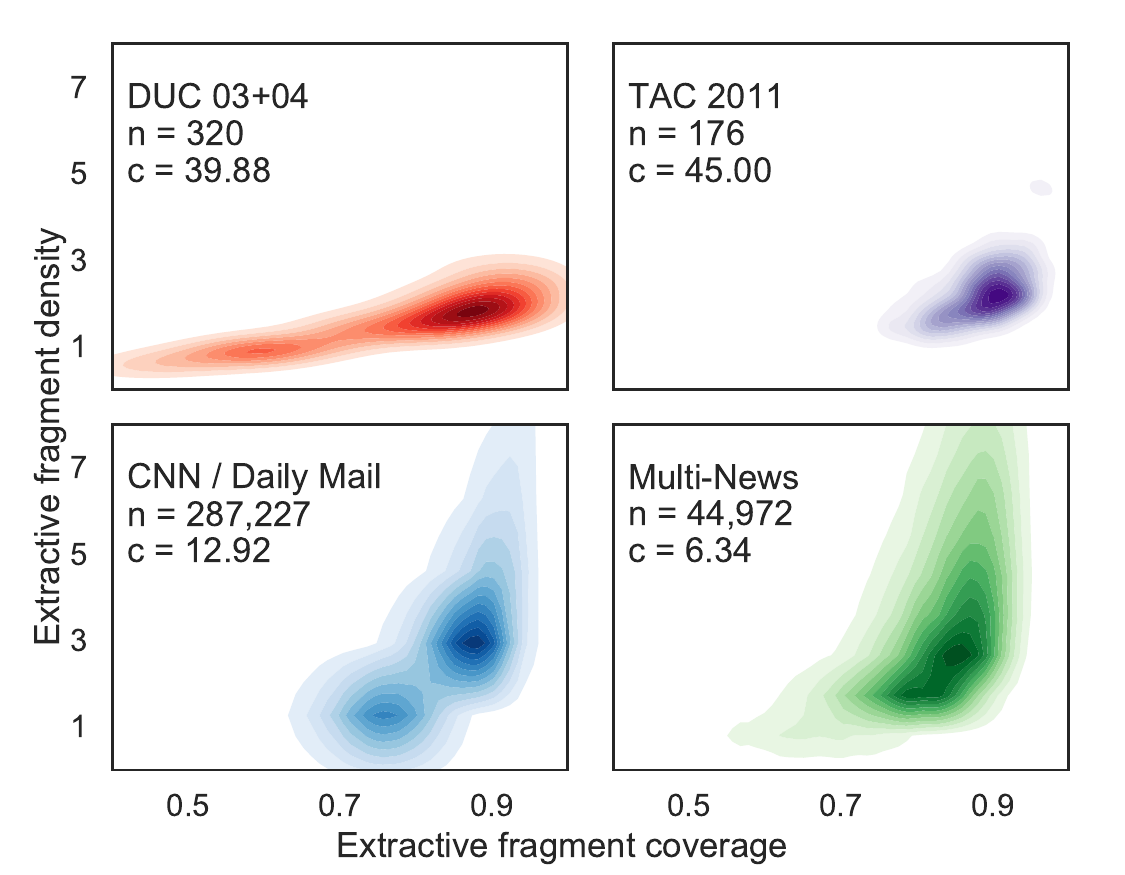}
    \caption{ Density estimation of extractive diversity scores as explained in Section \ref{diversity_section}. Large variability along the y-axis suggests variation in the average length of source sequences present in the summary, while the x axis shows variability in the average length of the extractive fragments to which summary words belong.
    }
    \label{fig:diversity}
    \vspace{-4mm}
\end{figure}

We report the percentage of n-grams in the gold summaries which do not appear in the input documents as a measure of how abstractive our summaries are in Table \ref{tab:ngrams}. As the table shows, the smaller MDS datasets tend to be more abstractive, but Multi-News is comparable and similar to the abstractiveness of SDS datasets. \newcite{Grusky:18} additionally define three measures of the extractive nature of a dataset, which we use here for a comparison. We extend these notions to the multi-document setting by concatenating the source documents and treating them as a single input. Extractive fragment coverage is the percentage of words in the summary that are from the source article, measuring the extent to which a summary is derivative of a text:
% \vspace{-2mm}
\begin{equation}
    \textit{COVERAGE(A,S)} = \frac{1}{|S|}\sum_{f \in F(A,S)} |f|
\end{equation}
% \vspace{-4mm}
%
where A is the article, S the summary, and $F(A,S)$ the set of all token sequences identified as extractive in a greedy manner; if there is a sequence of source tokens that is a prefix of the remainder of the summary, that is marked as extractive. Similarly, density is defined as the average length of the extractive fragment to which each summary word belongs: 
% \vspace{-2mm}
\begin{equation}
    \textit{DENSITY(A,S)} = \frac{1}{|S|}\sum_{f \in F(A,S)} |f|^2
\end{equation}
% \vspace{-4mm}
% 
Finally, compression ratio is defined as the word ratio between the articles and its summaries:
% \vspace{-2mm}
\begin{equation}
    \textit{COMPRESSION(A,S)} = \frac{|A|}{|S|}
\end{equation}
% \vspace{-4mm}
% 
These numbers are plotted using kernel density estimation in Figure \ref{fig:diversity}. As explained above, our summaries are larger on average, which corresponds to a lower compression rate. The variability along the x-axis (fragment coverage), suggests variability in the percentage of copied words, with the DUC data varying the most. In terms of y-axis (fragment density), our dataset shows variability in the average length of copied sequence, suggesting varying styles of word sequence arrangement. Our dataset exhibits extractive characteristics similar to the CNNDM dataset. 
   % Extractive diversity distribution and compression ratio, as introduced in \newcite{Grusky:18} and explained in \ref{diversity_section}, 

    % showing that examples across our dataset show larger variation in terms of extractive density and coverage compared to other MDS datasets and similar density to CNNDM.
    
% Irene + Tianwei + Alex
% \subsection{Proxy-A Distance}

% \begin{table}[h!]
% \centering
% \small
% \begin{tabular}{|ccc|} \hline
%           & CNNDM    & Multi-news \\ \hline
% DUC04      & 1.4279 & 1.4970   \\
% CNNDM   &      -    & 1.2564  \\

% \hline
% \end{tabular}
% \caption{Proxy-A Distance scores between datasets}
% \label{tab:proxy}
% \end{table}

% We also compute the Proxy-A Distance \cite{ganin2016domain} between the different datasets to show the discrepancy of them in Table \ref{tab:proxy}.  For each pair of the datasets, we first mix them up, and each document has the dataset name as the label,  leveraging a binary classification problem.  We extract TF-IDF features first and then choose the SVM classifier to do the classification,  and we marked the testing error to be $e$. Then the Proxy-A Distance score $PAD$ is given as:

% \begin{equation}
% \begin{aligned}
%     PAD = 2(1-2e)
% \end{aligned}
% \end{equation}

% A larger score means a larger discrepancy between the datasets. In Table \ref{tab:proxy}, we show that DUC 03 and DUC 04 datasets are quite similar. CNN and Daily mail are also similar pairs. However, these two datasets and DUC TAC groups have a relatively large gap. We also performed these calculations with respect to a sample of WikiSum dataset and found that dataset to be vastly different from the above datasets. These makes intuitive sense as Wikipedia articles differ largely in style in composition from news articles. 

% TODO should we mention duc01-03 and TAC?????
\subsection{Other Datasets}
As discussed above, large scale datasets for multi-document news summarization are lacking. There have been several attempts to create MDS datasets in other domains. \newcite{zopf18mds} introduce a multi-lingual MDS dataset based on English and German Wikipedia articles as summaries to create a set of about 7,000 examples. \newcite{liu18wikisum} use Wikipedia as well, creating a dataset of over two million examples. That paper uses Wikipedia references as input documents but largely relies on Google search to increase topic coverage. We, however, are focused on the news domain, and the source articles in our dataset are specifically cited by the corresponding summaries. Related work has also focused on opinion summarization in the multi-document setting; \newcite{angelidis18opinions} introduces a dataset of 600 Amazon product reviews.

\section{Preliminaries}
We introduce several common methods for summarization. 
\subsection{Pointer-generator Network}
% (Tianwei)

% one of the most commonly-used summarizer model at present. It combines both copying words from source documents and producing novel words from vocabulary, which allows accurate reproduction of information without sacrificing the ability to generate new words. 

% The model is a sequence-to-sequence attention model, including an encoder (a single-layer bi-LSTM) and a decoder (a single-layer LSTM) with attention mechanism. 

The pointer-generator network \cite{see2017ptr_gen} is a commonly-used encoder-decoder summarization model with attention \cite{bahdanau2014neural} which combines copying words from source documents and outputting words from a vocabulary.  The encoder converts each token $w_i$ in the document into the hidden state $h_i$. At each decoding step $t$, the decoder has a hidden state $d_t$. An attention distribution $a^t$ is calculated as in \cite{bahdanau2014neural} and is used to get the context vector $h_t^*$, which is a weighted sum of the encoder hidden states, representing the semantic meaning of the related document content for this decoding time step: 

\vspace{-4mm}
\begin{align}
\begin{split}
e_i^t &= v^T {\rm tanh} (W_h h_i + W_s d_t + b_{attn})  \\
a^t &= {\rm softmax}( e^t) \label{eq:pt} \\ 
h_{t}^{*} &= \sum_{i}a_{i}^{t}h_{i}^{t} 
\end{split}
\end{align}
The context vector $h_t^*$ and the decoder hidden state $d_t$ are then passed to two linear layers to produce the vocabulary distribution $P_{vocab}$. For each word, there is also a copy probability $P_{copy}$. It is the sum of the attention weights over all the word occurrences:

\vspace{-4mm}
\begin{align}
\begin{split}
    P_{vocab} &= {\rm softmax} (V^{'}(V[d_t, h_t^*] + b) + b^{'}) \\
    P_{copy} &= \sum_{i:w_i=w}a_i^t 
\end{split}
\end{align}
The pointer-generator network has a soft switch $p_{gen}$, which indicates whether to generate a word from vocabulary by sampling from $P_{vocab}$, or to copy a word from the source sequence by sampling from the copy probability $P_{copy}$. 

\vspace{-4mm}
\begin{equation}
      p_{gen} = \sigma (w_{h^*}^Th_t^* + w_d^Td_t + w_x^Tx_t + b_{ptr})
\end{equation}
where $x_t$ is the decoder input. The final probability distribution is a weighted sum of the vocabulary distribution and copy probability:

\begin{dmath} 
P(w) = p_{gen}P_{vocab}(w) + (1-p_{gen})P_{copy}(w)
\end{dmath} 

% \begin{gather}
%     % p_{gen} = \sum (w_{h^*}^T h^*_t + w_s^T s_t + w_x^T x_t + b_{ptr}) \\
%     % P(w) = p_{gen}P_{vocab}(w) + 
%     % (1-p_{gen})\textstyle\sum\nolimits_{w_i=w}a_i^t    
% \end{gather}

% The challanges when extended into MDS

\subsection{Transformer}
% (Tianwei)
The Transformer model replaces recurrent layers with self-attention in an encoder-decoder framework and has achieved state-of-the-art results in machine translation \cite{vaswani2017attention} and language modeling \cite{baevski2018adaptive, dai2019transformerxl}. The Transformer has also been successfully applied to SDS \cite{Gehrmann:18}. More specifically, for each word during encoding, the multi-head self-attention sub-layer allows the encoder to directly attend to all other words in a sentence in one step. Decoding contains the typical encoder-decoder attention mechanisms as well as self-attention to all previous generated output. The Transformer motivates the elimination of recurrence to allow more direct interaction among words in a sequence. 

% By eliminating the recurrence completely, the Transformer can avoid the long-term dependency problem of RNNs and also make it possible for parallel training.
% TODO also mention how we would like to incorporate our idea into the transformer
% shows a significantly better computational performance and accuracy in language understanding. Like most seq2seq models, the Transformer is also composed of an encoder and a decoder. But instead of using recurrence, the Transformer replace it with self-attention mechanism to model the dependencies in sequence. 

\subsection{MMR}
% (Irene)
% In MDS, we can apply a rank-based method to select sentences from the whole context as an extractive summary. 
Maximal Marginal Relevance (MMR) is an approach for combining query-relevance with information-novelty in the context of summarization \cite{carbonell1998use}. MMR produces a ranked list of the candidate sentences based on the relevance and redundancy to the query, which can be used to extract sentences. The score is calculated as follows: 

% \vspace{-5mm}

\begin{dmath} 
\label{eq:mmr}
{\text{MMR}}=\argmax_{ D_{ i }\in R\setminus S }  \left[ \lambda { \text{Sim} }_{ 1 } (D_{ i },Q)-(1-\lambda )\max _{ D_{ j }\in S }{ \text{Sim}_2 }(D_{ i },D_{ j } ) \right]
\end{dmath}
where $R$ is the collection of all candidate sentences, $Q$ is the query, $S$ is the set of sentences that have been selected, and $R \setminus  S$ is set of the un-selected ones. In general, each time we want to select a sentence, we have a ranking score for all the candidates that considers relevance and redundancy. 
A recent work \cite{lebanoff18mds} applied MMR for multi-document summarization by creating an external module and a supervised regression model for sentence importance. Our proposed method, however, incorporates MMR with the pointer-generator network in an end-to-end manner that learns parameters for similarity and redundancy. 

% . Parameters of MMR are learned during training without any feature engineering. 

% However, they did feature engineering when defining the similarity functions $\text{Sim}_1$ and $\text{Sim}_2$ and did not integrate totally with the pointer-generator network when they calculate MMR scores. 

\section{Hi-MAP Model}
\begin{figure}[t]
    \centering
    \includegraphics[width=\columnwidth]{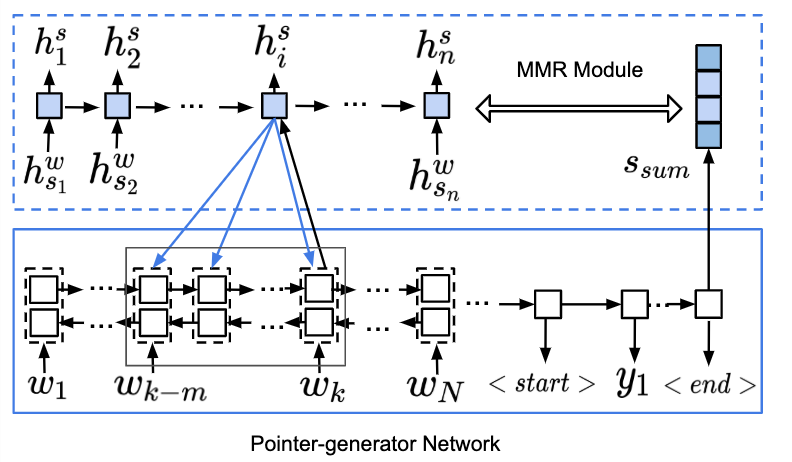}
    \caption{Our Hierarchical MMR-Attention Pointer-generator (Hi-MAP) model incorporates sentence-level representations and hidden-state-based MMR on top of a standard pointer-generator network.}
    \label{fig:mmr}
    % \vspace{-2mm}
\end{figure}

%(Alex+Irene) Model description
% TODO we need to describe models like PG transform, the difference between the opennmt and transformer implementations
In this section, we provide the details of our Hierarchical MMR-Attention Pointer-generator (Hi-MAP) model for  multi-document neural abstractive summarization. We expand the existing pointer-generator network model into a hierarchical network, which allows us to calculate sentence-level MMR scores. Our model consists of a pointer-generator network and an integrated MMR module, as shown in Figure \ref{fig:mmr}. 

% First we introduce our sentence representations, followed by  MMR-Attention and how it is integrated with the pointer-generator network.

% The pointer-generator network is a seq2seq model with word-level attention when generating a token at the decoding step. It is possible to concatenate all the documents tokens as the inputs to the pointer-generator network. However, one of the drawbacks of this approach is that it does not perform well when given a long document. Another drawback is for multi-document summarization, one needs to consider the redundancy when generating the summaries, which is a huge challenge for pointer-generator networks. 
% TODO explain rouge, random baselines etc

% The pointer-generator framework

\subsection{Sentence representations}
To expand our model into a hierarchical one, we compute sentence representations on both the encoder and decoder. The input is a collection of sentences $D=[s_1,s_2,..,s_n]$ from all the source documents, where a given sentence $s_i=[w_{k-m},w_{k-m+1},...,w_k]$ is made up of input word tokens. Word tokens from the whole document are treated as a single sequential input to a Bi-LSTM encoder as in the original encoder of the pointer-generator network from \newcite{see2017ptr_gen} (see bottom of Figure \ref{fig:mmr}). For each time step, the output of an input word token ${w_l}$ is $h^{w}_{l}$ (we use superscript $w$ to indicate word-level LSTM cells, $s$ for sentence-level).
% is the concatenation of the hidden states in two directions: $h^{w}_{l}=[\overleftarrow{h^{w}_{l}} ;\overrightarrow{h^{w}_{l}}]$ 

To obtain a representation for each sentence $s_i$, we take the encoder output of the last token for that sentence. If that token has an index of $k$ in the whole document $D$, then the sentence representation is marked as $h^{w}_{s_i}=h^{w}_{k}$. The word-level sentence embeddings of the document  $h^{w}_{D}=[h^{w}_{s_1},h^{w}_{s_2},..h^{w}_{s_n}]$ will be a sequence which is fed into a sentence-level LSTM network. Thus, for each input sentence $h^w_{s_i}$, we obtain an output hidden state $h^{s}_{s_i}$. We then get the final sentence-level embeddings $h^{s}_D = [h^{s}_{1},h^{s}_{2},..h^{s}_{n}]$ (we omit the subscript for sentences $s$). To obtain a summary representation, we simply treat the current decoded summary as a single sentence and take the output of the last step of the decoder: $s_{sum}$. We plan to investigate alternative methods for input and output sentence embeddings, such as separate LSTMs for each sentence, in future work.

% Note that, we did not make a sentence-level part on the decoding part, as we assume that the summary is a special long sentence and just keep it in word granularity. 

\subsection{MMR-Attention}

Now, we have all the sentence-level representation from both the articles and summary, and then we apply MMR to compute a ranking on the candidate sentences $h^{s}_D$. Intuitively, incorporating MMR will help determine salient sentences from the input at the current decoding step based on relevancy and redundancy. 

We follow Section 4.3 to compute MMR scores. Here, however, our query document is represented by the summary vector $s_{sum}$, and we want to rank the candidates in $h^{s}_D$. The MMR score for an input sentence $i$ is then defined as:

% And now we transfer the problem to be: given the reference, how can we rank the candidates in $h^s$? Or can we learn a distribution as a soft alignment between $h^s$ and $s_{sum}$? 

% $ h^{s}_ {s_i }\in h^{S}$ 
%{ {\text{MMR}} _{ h^{s}_ {s_i }} }
\begin{dmath} 
\label{eq:mmr-doc}
{ {\text{MMR}} _{i} } =  \lambda { \text{Sim} }_{ 1 } (h^{s}_ {i },s_{sum})-(1-\lambda )\max _{ {s_j} \in D, j \neq i }{ \text{Sim}_2 }(h^{s}_ {i },h^{s}_ {j } ) 
\end{dmath}
We then add a softmax function to normalize all the MMR scores of these candidates as a probability distribution.
\begin{dmath} 
\label{eq:mmr-doc-softmax}
\overline{{\text{MMR}}_{ i}}  = \frac{\exp({ {\text{MMR}} _{i} })}{\sum_{i}{\exp({ {\text{MMR}} _{ i} })}}
\end{dmath}
% In our experiment, we set $\lambda=0.5$.
Now we define the similarity function between each candidate sentence $h^s_{i}$ and summary sentence $s_{sum}$ to be:
\begin{equation}
    {\text{Sim}}_1 = {h^s_{i}}^T W_{\text{Sim}} s_{sum}
\end{equation}
where $W_{\text{Sim}}$ is a learned parameter used to transform $s_{sum}$ and ${h^s_{i}}$ into a common feature space. 
% this transformation aims to aline the feature spaces Since $s_{sum}$ and ${h^s_{s_i}}$ are from different feature space, we apply a transformation between the feature spaces using $W_{\text{Sim}}$.

For the second term of Equation \ref{eq:mmr-doc}, instead of choosing the maximum score from all candidates except for $h^s_{i}$, which is intended to find the candidate most similar to $h^s_{i}$, we choose to apply a self-attention model on $h^s_{i}$ and all the other candidates $h^s_{j}\in  h^{s}_D$. We then choose the largest weight as the final score:

\vspace{-4mm}
\begin{align} 
\begin{split}
\label{topic}
    v _ {ij} & = \tanh \left({ h^s_{j}}^T W _ {self}h^s_{i} \right) \\ 
    \beta _ { i j } & = \frac { \exp \left(  v _ { i j } \right) } { \sum _ { j } \exp \left(  v _ { i j } \right) } \\ 
    \text{score}_{i} & = \max_{j} (\beta_{i,j})  
\end{split}
\end{align}
Note that $W_{self}$ is also a trainable parameter. Eventually, the MMR score from Equation \ref{eq:mmr-doc} becomes:

\vspace{+2mm}
\begin{dmath} 
\label{eq:mmr-doc1}
{ {\text{MMR}} _{ i} } = \lambda { \text{Sim} }_{ 1 } (h^s_{i},s_{sum})-(1-\lambda ) \text{score}_{i}
\end{dmath}

\subsection{MMR-attention Pointer-generator}

After we calculate ${\overline{\text{MMR}}_{i}}$ for each sentence representation $h^s_{i}$, we use these scores to update the word-level attention weights for the pointer-generator model shown by the blue arrows in Figure \ref{fig:mmr}. Since $\overline{{\text{MMR}}_{i}}$ is a sentence weight for $h^s_{i}$, each token in the sentence will have the same value of $\overline{{\text{MMR}}_{i}}$. The new attention for each input token from Equation \ref{eq:pt} becomes: 
\begin{equation}
    \overline {a^{t}}=a^t \overline{{\text{MMR}}_i}    
\end{equation}
% Then we can effect the attention for each token via MMR scores by considering the sentence level information relevance and redundancy. 

\section{Experiments}
In this section we describe additional methods we compare with and present our assumptions and experimental process.

% \subsection{Baseline Methods}

% \textbf{Random:} We first concatenate all the sentences from each article in a document cluster and random choose one sentence as the system summary. For our dataset, \textit{Random-$k$} means we random take $k$ sentences from each source article and concatenate them as the summary.

% Since DUC data has about 10 sources for each event, if we take first one sentence from each, it is already very long. So we did not take first two or three sentences for DUC. 

% \textbf{PG-first} The method names start with \textit{PG} gives the baseline results for the pointer-generator network \cite{see2017ptr_gen}. We trained the original model on the CNN / Daily Mail first, then concatenated the first sentence from each article in that cluster as a single document and treated the decoded sentence as the summary of that cluster.

% \textbf{PG-concate} Similarly, we took the pre-trained original model with pointer-generator network but when decoding, we concatenated all contents from the articles.

\subsection{Baseline and Extractive Methods}
% TODO explain how we tried to use lexrank etc
\textbf{First} We concatenate the first sentence of each article in a document cluster as the system summary. For our dataset, \textit{First-$k$} means the first $k$ sentences from each source article will be concatenated as the summary. Due to the difference in gold summary length, we only use First-1 for DUC, as others would exceed the average summary length.

\noindent\textbf{LexRank} Initially proposed by \cite{erkan2004lexrank}, LexRank is a graph-based method for computing relative importance in extractive summarization.

\noindent\textbf{TextRank} Introduced by \cite{mihalcea2004textrank}, TextRank is a graph-based ranking model.  Sentence importance scores are computed based on eigenvector centrality within a global graph from the corpus.

\noindent\textbf{MMR} In addition to incorporating MMR in our pointer generator network, we use this original method as an extractive summarization baseline.
When testing on DUC data, we set these extractive methods to give an output of 100 tokens and 300 tokens for Multi-News data.
% \vspace{-2mm}
\subsection{Neural Abstractive Methods}
% pre-processing, truncated.. (Alex)
% \vspace{-2mm}
\textbf{PG-Original, PG-MMR} These are the original pointer-generator network models reported by \cite{lebanoff18mds}.

\noindent\textbf{PG-BRNN} The PG-BRNN model is a pointer-generator implementation from OpenNMT\footnote{\url{https://github.com/OpenNMT/OpenNMT-py/blob/master/docs/source/Summarization.md}}. As in the original paper \cite{see2017ptr_gen}, we use a 1-layer bi-LSTM as encoder, with 128-dimensional word-embeddings and 256-dimensional hidden states for each direction. The decoder is a 512-dimensional single-layer LSTM. We include this for reference in addition to PG-Original, as our Hi-MAP code builds upon this implementation. 

\noindent\textbf{CopyTransformer} Instead of using an LSTM, the CopyTransformer model used in \newcite{Gehrmann:18} uses a 4-layer Transformer of 512 dimensions for encoder and decoder.  One of the attention heads is chosen randomly as the copy distribution. This model and the PG-BRNN are run without the bottom-up masked attention for inference from \newcite{Gehrmann:18} as we did not find a large improvement when reproducing the model on this data. 

% We did not use the bottom-up masked attention during inference as in \newcite{Gehrmann:18} due to difficulties in reproducing results from the original paper. 
% \noindent\textbf{Hi-MAP} We set the pointer-generator setting to be the same with PG-BRNN. Our sentence-level encoder is a 1-layer bidirectional LSTM network, with the hidden state dimension being 256 in each direction. The sentence representation dimension is also 256. 
% \vspace{-2mm}
\subsection{Experimental Setting}

Following the setting from \cite{lebanoff18mds}, we report ROUGE \cite{lin2004rouge} scores, which measure the overlap of unigrams (R-1), bigrams (R-2) and skip bigrams with a max distance of four words (R-SU).  For the neural abstractive models, we truncate input articles to 500 tokens in the following way: for each example with $S$ source input documents, we take the first 500$/S$ tokens from each source document.  As some source documents may be shorter, we iteratively determine the number of tokens to take from each document until the 500 token quota is reached. Having determined the number of tokens per source document to use, we concatenate the truncated source documents into a single mega-document. This effectively reduces MDS to SDS on longer documents, a commonly-used assumption for recent neural MDS papers \cite{17classification,liu18wikisum,lebanoff18mds}. We chose 500 as our truncation size as related MDS work did not find significant improvement when increasing input length from 500 to 1000 tokens \cite{liu18wikisum}. We simply introduce a special token between source documents to aid our models in detecting document-to-document relationships and leave direct modeling of this relationship, as well as modeling longer input sequences, to future work. We hope that the dataset we introduce will promote such work. For our Hi-MAP model, we applied a 1-layer bidirectional LSTM network, with the hidden state dimension 256 in each direction. The sentence representation dimension is also 256. We set the $\lambda=0.5$ to calculate the MMR value in Equation \ref{eq:mmr-doc}.

\begin{table}[h!]
\centering
\small
\resizebox{\columnwidth}{!}{\begin{tabular}{|l|ccc|}
\hline
    \textbf{Method}   & \textbf{R-1}    & \textbf{R-2}    & \textbf{R-SU}    \\ \hline \hline
% Random & 11.8 & 1.73 & 1.27 \\
First  & 30.77 & 8.27 & 7.35 \\ \hline

% PG-first	&	22.86	&	4.66	&	5.46	\\
% PG-concate	&	28.98	&	6.67	&	8.03	\\

% \hline

LexRank {\tiny \cite{erkan2004lexrank}}	&	35.56	&	7.87	&	11.86	\\
TextRank {\tiny \cite{mihalcea2004textrank}}	&	33.16	&	6.13	&	10.16	\\
MMR	{\tiny \cite{carbonell1998use}} &	30.14	&	4.55	&	8.16	\\
\hline

PG-Original{\tiny\cite{lebanoff18mds}} & 31.43  &  6.03  & 10.01 \\
PG-MMR{\tiny\cite{lebanoff18mds}} & \textbf{36.42} & \textbf{9.36} & \textbf{13.23} \\ 
\hline

PG-BRNN {\tiny \cite{Gehrmann:18} }	&	29.47	&	6.77	&	7.56	\\
CopyTransformer {\tiny \cite{Gehrmann:18} }	&	28.54	&	6.38	&	7.22	\\
\hline
Hi-MAP (Our Model)  & 35.78 & 8.90 & 11.43 \\

\hline
\end{tabular}}
\caption{ROUGE scores on the DUC 2004 dataset for models trained on CNNDM data, as in \newcite{lebanoff18mds}.\footnotemark}
\label{tab:duc_baseline}
\end{table}

\footnotetext{As our focus was on deep methods for MDS, we only tested several non-neural baselines. However, other classical methods deserve more attention, for which we refer the reader to \newcite{Hong14} and leave the implementation of these methods on Multi-News for future work.}

\begin{table}[h!]
\centering
\small
\resizebox{\columnwidth}{!}{\begin{tabular}{|l|ccc|}
\hline
    \textbf{Method}   & \textbf{R-1}    & \textbf{R-2}    & \textbf{R-SU}    \\ \hline \hline
% Random-1	&	20.97	&	5.12	&	3.83	\\
% Random-2	&	31.62	&	8.14	&	8.94	\\
% Random-3	&	36.79	&	9.92	&	12.26	\\
First-1	&	26.83	&	7.25	&	6.46	\\
First-2	&	35.99	&	10.17	&	12.06	\\
First-3	&	39.41	&	11.77	&	14.51	\\
\hline

% PG-first   & TBD & TBD & 0.TBD \\
% PG-concate  & TBD & TBD & 0.TBD \\ \hline

LexRank {\tiny \cite{erkan2004lexrank}}	&	38.27	&	12.70	&	13.20	\\
TextRank {\tiny \cite{mihalcea2004textrank}}	&	38.44	&	13.10	&	13.50	\\
MMR {\tiny \cite{carbonell1998use}}	&	38.77	&	11.98	&	12.91	\\

\hline

PG-Original {\tiny\cite{lebanoff18mds}} & 41.85 & 12.91	& 16.46 \\
PG-MMR {\tiny \cite{lebanoff18mds}} & 40.55 & 12.36 & 15.87 \\ 

\hline

PG-BRNN {\tiny \cite{Gehrmann:18} } & 42.80 & 14.19 & 16.75 \\
CopyTransformer {\tiny \cite{Gehrmann:18} }  & \textbf{43.57} & 14.03 & 17.37 \\
\hline

Hi-MAP (Our Model)  & 43.47 & \textbf{14.89} & \textbf{17.41} \\
\hline
\end{tabular}}
\caption{ROUGE scores for models trained and tested on the Multi-News dataset.}
\label{tab:multinews_baseline}
\end{table}

\vspace{-2mm}
\section{Analysis and Discussion}

% TODO need to explain that we train on CNNDM and test on DUC
In Table \ref{tab:duc_baseline} and Table \ref{tab:multinews_baseline} we report ROUGE scores on DUC 2004 and Multi-News datasets respectively.
We use DUC 2004, as results on this dataset are reported in \newcite{lebanoff18mds}, although this dataset is not the focus of this work. For results on DUC 2004, models were trained on the CNNDM dataset, as in \newcite{lebanoff18mds}. PG-BRNN and CopyTransformer models, which were pretrained by OpenNMT on CNNDM, were applied to DUC without additional training, analogous to PG-Original. We also experimented with training on Multi-News and testing on DUC data, but we did not see significant improvements. We attribute the generally low performance of pointer-generator, CopyTransformer and Hi-MAP to domain differences between DUC and CNNDM as well as DUC and Multi-News. These domain differences are evident in the statistics and extractive metrics discussed in Section 3.

\begin{table}[t!]
\centering
\small
\resizebox{\columnwidth}{!}{\begin{tabular}{c c c c  } \Xhline{2\arrayrulewidth}
             \textbf{Method} & \textbf{Informativeness} & \textbf{Fluency} & \textbf{Non-Redundancy} \\ \Xhline{\arrayrulewidth}
             PG-MMR& 95 & 70 & 45   \\ %\Xhline{2\arrayrulewidth}
             Hi-MAP & 85 & 75 & 100   \\ %\Xhline{2\arrayrulewidth}
             CopyTransformer & 99 & 100 & 107   \\
             Human & 150 & 150 & 149   \\
             
             \Xhline{2\arrayrulewidth}
\end{tabular}}
\caption{Number of times a system was chosen as best in pairwise comparisons according to informativeness, fluency and non-redundancy.}
\label{tab:evaluation}
\vspace{-3mm}
\end{table}

Additionally, for both DUC and Multi-News testing, we experimented with using the output of 500 tokens from extractive methods (LexRank, TextRank and MMR) as input to the abstractive model. However, this did not improve results. We believe this is because our truncated input mirrors the First-3 baseline, which outperforms these three extractive methods and thus may provide more information as input to the abstractive model.

Our model outperforms PG-MMR when trained and tested on the Multi-News dataset. We see much-improved model performances when trained and tested on in-domain Multi-News data. The Transformer performs best in terms of R-1 while Hi-MAP outperforms it on R-2 and R-SU. Also, we notice a drop in performance between PG-original, and PG-MMR (which takes the pre-trained PG-original and applies MMR on top of the model). Our PG-MMR results correspond to \textit{PG-MMR w\/ Cosine} reported in \newcite{lebanoff18mds}. 
We trained their sentence regression model on Multi-News data and leave the investigation of transferring regression models from SDS to Multi-News for future work.

% TODO explain why results from MDS code are worse
In addition to automatic evaluation, we performed human evaluation to compare the summaries produced. We used Best-Worst Scaling \cite{louvrier1991, louvrier2015}, which has shown to be more reliable than rating scales \cite{kiritchenko17} and has been used to evaluate  summaries \cite{narayan18xsum, angelidis18opinions}. Annotators were presented with the same input that the systems saw at testing time; input documents were truncated, and we separated input documents by visible spaces in our annotator interface. We chose three native English speakers as annotators. They were presented with input documents, and summaries generated by two out of four systems, and were asked to determine which summary was better and which was worse in terms of \textit{informativeness} (is the meaning in the input text preserved in the summary?), \textit{fluency} (is the summary written in well-formed and grammatical English?) and 
\textit{non-redundancy} (does the summary avoid repeating information?). We randomly selected 50 documents from the Multi-News test set and compared all possible combinations of two out of four systems. We chose to compare PG-MMR, CopyTransformer, Hi-MAP and gold summaries. The order of summaries was randomized per example.

The results of our pairwise human-annotated comparison are shown in Table \ref{tab:evaluation}. Human-written summaries were easily marked as better than other systems, which, while expected, shows that there is much room for improvement in producing readable, informative summaries. We performed pairwise comparison of the models over the three metrics combined, using a one-way ANOVA with Tukey HSD tests and $p$ value of 0.05. Overall, statistically significant differences were found between human summaries score and all other systems, CopyTransformer and the other two models, and our Hi-MAP model compared to PG-MMR. Our Hi-MAP model performs comparably to PG-MMR on informativeness and fluency but much better in terms of non-redundancy. We believe that the incorporation of learned parameters for similarity and redundancy reduces redundancy in our output summaries. In future work, we would like to incorporate MMR into Transformer models to benefit from their fluent summaries. 
% The CopyTransformer is better rated in terms of informativeness and non-redundancy in a statistically significant manner. 
% (using a one-way ANOVA < 0.01)
\section{Conclusion}
\vspace{-2mm}
In this paper we introduce Multi-News, the first large-scale multi-document news summarization dataset. We hope that this dataset will promote work in multi-document summarization similar to the progress seen in the single-document case. Additionally, we introduce an end-to-end model which incorporates MMR into a pointer-generator network, which performs competitively compared to previous multi-document summarization models. We also benchmark methods on our dataset. In the future we plan to explore interactions among documents beyond concatenation and experiment with summarizing longer input documents.

% to promote further work in large-scale MDS news summarization.

% 
% \section*{Acknowledgments}

% The acknowledgments should go immediately before the references.  Do
% not number the acknowledgments section. Do not include this section
% when submitting your paper for review. \\
% \clearpage
\bibliography{acl2019}

\begin{thebibliography}{46}
\expandafter\ifx\csname natexlab\endcsname\relax\def\natexlab#1{#1}\fi

\bibitem[{NYT(2008)}]{NYT}
 2008.
\newblock \href {https://catalog.ldc.upenn.edu/LDC2008T19} {The {N}ew {Y}ork
  {T}imes {A}nnotated {C}orpus}.

\bibitem[{Angelidis and Lapata(2018)}]{angelidis18opinions}
Stefanos Angelidis and Mirella Lapata. 2018.
\newblock Summarizing {O}pinions: {A}spect {E}xtraction {M}eets {S}entiment
  {P}rediction and {T}hey {A}re {B}oth {W}eakly {S}upervised.
\newblock In \emph{Proceedings of the 2018 Conference on Empirical Methods in
  Natural Language Processing, Brussels, Belgium, October 31 - November 4,
  2018}, pages 3675--3686.

\bibitem[{Baevski and Auli(2019)}]{baevski2018adaptive}
Alexei Baevski and Michael Auli. 2019.
\newblock Adaptive input representations for neural language modeling.
\newblock In \emph{International Conference on Learning Representations}.

\bibitem[{Bahdanau et~al.(2014)Bahdanau, Cho, and Bengio}]{bahdanau2014neural}
Dzmitry Bahdanau, Kyunghyun Cho, and Yoshua Bengio. 2014.
\newblock Neural {M}achine {T}ranslation by {J}ointly {L}earning to {A}lign and
  {T}ranslate.
\newblock \emph{arXiv preprint arXiv:1409.0473}.

\bibitem[{Barzilay et~al.(1999)Barzilay, McKeown, and
  Elhadad}]{barzilay99fusion}
Regina Barzilay, Kathleen~R. McKeown, and Michael Elhadad. 1999.
\newblock Information {F}usion in the {C}ontext of {M}ulti-{D}ocument
  {S}ummarization.
\newblock In \emph{27th Annual Meeting of the Association for Computational
  Linguistics, University of Maryland, College Park, Maryland, USA, 20-26 June
  1999.}

\bibitem[{Baumel et~al.(2018)Baumel, Eyal, and Elhadad}]{baumel18mds}
Tal Baumel, Matan Eyal, and Michael Elhadad. 2018.
\newblock Query {F}ocused {A}bstractive {S}ummarization: {I}ncorporating
  {Q}uery {R}elevance, {M}ulti-{D}ocument {C}overage, and {S}ummary {L}ength
  {C}onstraints into seq2seq {M}odels.
\newblock \emph{CoRR}, abs/1801.07704.

\bibitem[{Cao et~al.(2017)Cao, Li, Li, and Wei}]{17classification}
Ziqiang Cao, Wenjie Li, Sujian Li, and Furu Wei. 2017.
\newblock Improving {M}ulti-{D}ocument {S}ummarization via {T}ext
  {C}lassification.
\newblock In \emph{Proceedings of the Thirty-First {AAAI} Conference on
  Artificial Intelligence, February 4-9, 2017, San Francisco, California,
  {USA.}}, pages 3053--3059.

\bibitem[{Carbonell and Goldstein(1998)}]{carbonell1998use}
Jaime Carbonell and Jade Goldstein. 1998.
\newblock The {U}se of {MMR}, {D}iversity-{B}ased {R}eranking for {R}eordering
  {D}ocuments and {P}roducing {S}ummaries.
\newblock In \emph{Proceedings of the 21st annual international ACM SIGIR
  conference on Research and development in information retrieval}, pages
  335--336. ACM.

\bibitem[{{\c{C}}elikyilmaz et~al.(2018){\c{C}}elikyilmaz, Bosselut, He, and
  Choi}]{celikyilmaz18rl}
Asli {\c{C}}elikyilmaz, Antoine Bosselut, Xiaodong He, and Yejin Choi. 2018.
\newblock Deep {C}ommunicating {A}gents for {A}bstractive {S}ummarization.
\newblock In \emph{Proceedings of the 2018 Conference of the North American
  Chapter of the Association for Computational Linguistics: Human Language
  Technologies, {NAACL-HLT} 2018, New Orleans, Louisiana, USA, June 1-6, 2018,
  Volume 1 (Long Papers)}, pages 1662--1675.

\bibitem[{Cheng and Lapata(2016)}]{cheng16ext}
Jianpeng Cheng and Mirella Lapata. 2016.
\newblock Neural {S}ummarization by {E}xtracting {S}entences and {W}ords.
\newblock In \emph{Proceedings of the 54th Annual Meeting of the Association
  for Computational Linguistics, {ACL} 2016, August 7-12, 2016, Berlin,
  Germany, Volume 1: Long Papers}.

\bibitem[{Chopra et~al.(2016)Chopra, Auli, and Rush}]{chopra16og}
Sumit Chopra, Michael Auli, and Alexander~M. Rush. 2016.
\newblock Abstractive {S}entence {S}ummarization with {A}ttentive {R}ecurrent
  {N}eural {N}etworks.
\newblock In \emph{{NAACL} {HLT} 2016, The 2016 Conference of the North
  American Chapter of the Association for Computational Linguistics: Human
  Language Technologies, San Diego California, USA, June 12-17, 2016}, pages
  93--98.

\bibitem[{Chu and Liu(2019)}]{liu18unsupervised}
Eric Chu and Peter Liu. 2019.
\newblock \href {http://proceedings.mlr.press/v97/chu19b.html} {{M}ean{S}um: A
  neural model for unsupervised multi-document abstractive summarization}.
\newblock In \emph{Proceedings of the 36th International Conference on Machine
  Learning}, volume~97 of \emph{Proceedings of Machine Learning Research},
  pages 1223--1232, Long Beach, California, USA. PMLR.

\bibitem[{Cohan et~al.(2018)Cohan, Dernoncourt, Kim, Bui, Kim, Chang, and
  Goharian}]{cohan18discourse}
Arman Cohan, Franck Dernoncourt, Doo~Soon Kim, Trung Bui, Seokhwan Kim, Walter
  Chang, and Nazli Goharian. 2018.
\newblock A {D}iscourse-{A}ware {A}ttention {M}odel for {A}bstractive
  {S}ummarization of {L}ong {D}ocuments.
\newblock In \emph{Proceedings of the 2018 Conference of the North American
  Chapter of the Association for Computational Linguistics: Human Language
  Technologies, NAACL-HLT, New Orleans, Louisiana, USA, June 1-6, 2018, Volume
  2 (Short Papers)}, pages 615--621.

\bibitem[{Dai et~al.(2019)Dai, Yang, Yang, Cohen, Carbonell, Le, and
  Salakhutdinov}]{dai2019transformerxl}
Zihang Dai, Zhilin Yang, Yiming Yang, William~W. Cohen, Jaime Carbonell,
  Quoc~V. Le, and Ruslan Salakhutdinov. 2019.
\newblock Transformer-{XL}: Language modeling with longer-term dependency.

\bibitem[{Erkan and Radev(2004)}]{erkan2004lexrank}
G{\"u}nes Erkan and Dragomir~R Radev. 2004.
\newblock Lexrank: {G}raph-{B}ased {L}exical {C}entrality as {S}alience in
  {T}ext {S}ummarization.
\newblock \emph{Journal of artificial intelligence research}, 22:457--479.

\bibitem[{Ganesan et~al.(2010)Ganesan, Zhai, and Han}]{ganesan10opinosis}
Kavita Ganesan, ChengXiang Zhai, and Jiawei Han. 2010.
\newblock Opinosis: {A} {G}raph {B}ased {A}pproach to {A}bstractive
  {S}ummarization of {H}ighly {R}edundant {O}pinions.
\newblock In \emph{{COLING} 2010, 23rd International Conference on
  Computational Linguistics, Proceedings of the Conference, 23-27 August 2010,
  Beijing, China}, pages 340--348.

\bibitem[{Gehrmann et~al.(2018)Gehrmann, Deng, and Rush}]{Gehrmann:18}
Sebastian Gehrmann, Yuntian Deng, and Alexander~M. Rush. 2018.
\newblock Bottom-{U}p {A}bstractive {S}ummarization.
\newblock In \emph{Proceedings of the 2018 Conference on Empirical Methods in
  Natural Language Processing, Brussels, Belgium, October 31 - November 4,
  2018}, pages 4098--4109.

\bibitem[{Grusky et~al.(2018)Grusky, Naaman, and Artzi}]{Grusky:18}
Max Grusky, Mor Naaman, and Yoav Artzi. 2018.
\newblock Newsroom: {A} {D}ataset of 1.3 {M}illion {S}ummaries with {D}iverse
  {E}xtractive {S}trategies.
\newblock \emph{CoRR}, abs/1804.11283.

\bibitem[{Haghighi and Vanderwende(2009)}]{haghighi09content}
Aria Haghighi and Lucy Vanderwende. 2009.
\newblock Exploring {C}ontent {M}odels for {M}ulti-{D}ocument {S}ummarization.
\newblock In \emph{Human Language Technologies: Conference of the North
  American Chapter of the Association of Computational Linguistics,
  Proceedings, May 31 - June 5, 2009, Boulder, Colorado, {USA}}, pages
  362--370.

\bibitem[{Hermann et~al.(2015)Hermann, Kocisk{\'{y}}, Grefenstette, Espeholt,
  Kay, Suleyman, and Blunsom}]{Hermann:15}
Karl~Moritz Hermann, Tom{\'{a}}s Kocisk{\'{y}}, Edward Grefenstette, Lasse
  Espeholt, Will Kay, Mustafa Suleyman, and Phil Blunsom. 2015.
\newblock Teaching {M}achines to {R}ead and {C}omprehend.
\newblock In \emph{Advances in Neural Information Processing Systems 28: Annual
  Conference on Neural Information Processing Systems 2015, December 7-12,
  2015, Montreal, Quebec, Canada}, pages 1693--1701.

\bibitem[{Hong et~al.(2014)Hong, Conroy, Favre, Kulesza, Lin, and
  Nenkova}]{Hong14}
Kai Hong, John~M. Conroy, Beno{\^{\i}}t Favre, Alex Kulesza, Hui Lin, and Ani
  Nenkova. 2014.
\newblock A repository of state of the art and competitive baseline summaries
  for generic news summarization.
\newblock In \emph{Proceedings of the Ninth International Conference on
  Language Resources and Evaluation, {LREC} 2014, Reykjavik, Iceland, May
  26-31, 2014.}, pages 1608--1616.

\bibitem[{Kiritchenko and Mohammad(2017)}]{kiritchenko17}
Svetlana Kiritchenko and Saif Mohammad. 2017.
\newblock Best-{W}orst {S}caling {M}ore {R}eliable than {R}ating {S}cales: {A}
  {C}ase {S}tudy on {S}entiment {I}ntensity {A}nnotation.
\newblock In \emph{Proceedings of the 55th Annual Meeting of the Association
  for Computational Linguistics, {ACL} 2017, Vancouver, Canada, July 30 -
  August 4, Volume 2: Short Papers}, pages 465--470.

\bibitem[{Lebanoff et~al.(2018)Lebanoff, Song, and Liu}]{lebanoff18mds}
Logan Lebanoff, Kaiqiang Song, and Fei Liu. 2018.
\newblock Adapting the {N}eural {E}ncoder-{D}ecoder {F}ramework from {S}ingle
  to {M}ulti-{D}ocument summarization.
\newblock In \emph{Proceedings of the 2018 Conference on Empirical Methods in
  Natural Language Processing, Brussels, Belgium, October 31 - November 4,
  2018}, pages 4131--4141.

\bibitem[{Lin(2004)}]{lin2004rouge}
Chin-Yew Lin. 2004.
\newblock {R}ouge: {A} {P}ackage for {A}utomatic {E}valuation of {S}ummaries.
\newblock \emph{Text Summarization Branches Out}.

\bibitem[{Liu et~al.(2018)Liu, Saleh, Pot, Goodrich, Sepassi, Kaiser, and
  Shazeer}]{liu18wikisum}
Peter~J. Liu, Mohammad Saleh, Etienne Pot, Ben Goodrich, Ryan Sepassi, Lukasz
  Kaiser, and Noam Shazeer. 2018.
\newblock Generating {W}ikipedia by {S}ummarizing {L}ong {S}equences.
\newblock \emph{CoRR}, abs/1801.10198.

\bibitem[{Louviere et~al.(2015)Louviere, Flynn, and Marley}]{louvrier2015}
Jordan Louviere, Terry Flynn, and A.~A.~J. Marley. 2015.
\newblock \emph{Best-{W}orst {S}caling: {T}heory, {M}ethods and
  {A}pplications}.

\bibitem[{Louviere and Woodworth(1991)}]{louvrier1991}
Jordan~J Louviere and George~G Woodworth. 1991.
\newblock \emph{Best-{W}orst {S}caling: {A} {M}odel for the {L}argest
  {D}ifference {J}udgments.}

\bibitem[{McKeown and Radev(1995)}]{McKeown&Radev95}
Kathleen~R. McKeown and Dragomir~R. Radev. 1995.
\newblock Generating summaries of multiple news articles.
\newblock In \emph{Proceedings, ACM Conference on Research and Development in
  Information Retrieval SIGIR'95}, pages 74--82, Seattle, Washington.

\bibitem[{Mihalcea and Tarau(2004)}]{mihalcea2004textrank}
Rada Mihalcea and Paul Tarau. 2004.
\newblock Textrank: {B}ringing {O}rder into {T}ext.
\newblock In \emph{Proceedings of the 2004 conference on empirical methods in
  natural language processing}.

\bibitem[{Nallapati et~al.(2016{\natexlab{a}})Nallapati, Zhou, and
  Ma}]{nallapai16b}
Ramesh Nallapati, Bowen Zhou, and Mingbo Ma. 2016{\natexlab{a}}.
\newblock Classify or {S}elect: {N}eural {A}rchitectures for {E}xtractive
  {D}ocument {S}ummarization.
\newblock \emph{CoRR}, abs/1611.04244.

\bibitem[{Nallapati et~al.(2016{\natexlab{b}})Nallapati, Zhou, dos Santos,
  G{\"{u}}l{\c{c}}ehre, and Xiang}]{nallapati16a}
Ramesh Nallapati, Bowen Zhou, C{\'{\i}}cero~Nogueira dos Santos, {\c{C}}aglar
  G{\"{u}}l{\c{c}}ehre, and Bing Xiang. 2016{\natexlab{b}}.
\newblock Abstractive {T}ext {S}ummarization {U}sing {S}equence-to-{S}equence
  {RNN}s and {B}eyond.
\newblock In \emph{Proceedings of the 20th {SIGNLL} Conference on Computational
  Natural Language Learning, CoNLL 2016, Berlin, Germany, August 11-12, 2016},
  pages 280--290.

\bibitem[{Napoles et~al.(2012)Napoles, Gormley, and Durme}]{napoles12giga}
Courtney Napoles, Matthew~R. Gormley, and Benjamin~Van Durme. 2012.
\newblock Annotated {G}igaword.
\newblock In \emph{Proceedings of the Joint Workshop on Automatic Knowledge
  Base Construction and Web-scale Knowledge Extraction, AKBC-WEKEX@NAACL-HLT
  2012, Montr{\`{e}}al, Canada, June 7-8, 2012}, pages 95--100.

\bibitem[{Narayan et~al.(2018{\natexlab{a}})Narayan, Cohen, and
  Lapata}]{narayan18xsum}
Shashi Narayan, Shay~B. Cohen, and Mirella Lapata. 2018{\natexlab{a}}.
\newblock Don't {G}ive {M}e the {D}etails, {J}ust the {S}ummary!
  {T}opic-{A}ware {C}onvolutional {N}eural {N}etworks for {E}xtreme
  {S}ummarization.
\newblock In \emph{Proceedings of the 2018 Conference on Empirical Methods in
  Natural Language Processing}, pages 1797--1807. Association for Computational
  Linguistics.

\bibitem[{Narayan et~al.(2018{\natexlab{b}})Narayan, Cohen, and
  Lapata}]{narayan18rl}
Shashi Narayan, Shay~B. Cohen, and Mirella Lapata. 2018{\natexlab{b}}.
\newblock Ranking {S}entences for {E}xtractive {S}ummarization with
  {R}einforcement {L}earning.
\newblock In \emph{Proceedings of the 2018 Conference of the North American
  Chapter of the Association for Computational Linguistics: Human Language
  Technologies, {NAACL-HLT} 2018, New Orleans, Louisiana, USA, June 1-6, 2018,
  Volume 1 (Long Papers)}, pages 1747--1759.

\bibitem[{Owczarzak and Dang(2011)}]{owczarzak2011}
Karolina Owczarzak and Hoa~Trang Dang. 2011.
\newblock Overview of the {TAC} 2011 {S}ummarization {T}rack: {G}uided {T}ask
  and {AESOP} {T}ask.

\bibitem[{Paul and James(2004)}]{paul2004introduction}
Over Paul and Yen James. 2004.
\newblock An {I}ntroduction to {DUC}-2004.
\newblock In \emph{Proceedings of the 4th Document Understanding Conference
  (DUC 2004)}.

\bibitem[{Paulus et~al.(2017)Paulus, Xiong, and Socher}]{paulus17rl}
Romain Paulus, Caiming Xiong, and Richard Socher. 2017.
\newblock A {D}eep {R}einforced {M}odel for {A}bstractive {S}ummarization.
\newblock \emph{CoRR}, abs/1705.04304.

\bibitem[{Radev et~al.(2000)Radev, Jing, and Budzikowska}]{radev00centroid}
Dragomir~R. Radev, Hongyan Jing, and Malgorzata Budzikowska. 2000.
\newblock Centroid-{B}ased {S}ummarization of {M}ultiple {D}ocuments:
  {S}entence {E}xtraction utility-based evaluation, and user studies.
\newblock \emph{CoRR}, cs.CL/0005020.

\bibitem[{Radev and McKeown(1998)}]{radev98mds}
Dragomir~R. Radev and Kathleen~R. McKeown. 1998.
\newblock Generating {N}atural {L}anguage {S}ummaries from {M}ultiple
  {O}n-{L}ine {S}ources.
\newblock \emph{Computational Linguistics}, 24(3):469--500.

\bibitem[{Rush et~al.(2015)Rush, Chopra, and Weston}]{Rush:15}
Alexander~M. Rush, Sumit Chopra, and Jason Weston. 2015.
\newblock A {N}eural {A}ttention {M}odel for {A}bstractive {S}entence
  {S}ummarization.
\newblock In \emph{Proceedings of the 2015 Conference on Empirical Methods in
  Natural Language Processing, {EMNLP} 2015, Lisbon, Portugal, September 17-21,
  2015}, pages 379--389.

\bibitem[{See et~al.(2017)See, Liu, and Manning}]{see2017ptr_gen}
Abigail See, Peter~J Liu, and Christopher~D Manning. 2017.
\newblock Get {T}o {T}he {P}oint: {S}ummarization with {P}ointer-{G}enerator
  {N}etworks.
\newblock In \emph{Proceedings of the 55th Annual Meeting of the Association
  for Computational Linguistics (Volume 1: Long Papers)}, volume~1, pages
  1073--1083.

\bibitem[{Vaswani et~al.(2017)Vaswani, Shazeer, Parmar, Uszkoreit, Jones,
  Gomez, Kaiser, and Polosukhin}]{vaswani2017attention}
Ashish Vaswani, Noam Shazeer, Niki Parmar, Jakob Uszkoreit, Llion Jones,
  Aidan~N Gomez, {\L}ukasz Kaiser, and Illia Polosukhin. 2017.
\newblock Attention is all you need.
\newblock In \emph{Advances in Neural Information Processing Systems}, pages
  5998--6008.

\bibitem[{Yasunaga et~al.(2017)Yasunaga, Zhang, Meelu, Pareek, Srinivasan, and
  Radev}]{yasunaga17graph}
Michihiro Yasunaga, Rui Zhang, Kshitijh Meelu, Ayush Pareek, Krishnan
  Srinivasan, and Dragomir~R. Radev. 2017.
\newblock Graph-{B}ased {N}eural {M}ulti-{D}ocument {S}ummarization.
\newblock In \emph{Proceedings of CoNLL-2017}. Association for Computational
  Linguistics.

\bibitem[{Zhang et~al.(2018{\natexlab{a}})Zhang, Tan, and Wan}]{zhang18mds}
Jianmin Zhang, Jiwei Tan, and Xiaojun Wan. 2018{\natexlab{a}}.
\newblock Adapting {N}eural {S}ingle-{D}ocument {S}ummarization {M}odel for
  {A}bstractive {M}ulti-{D}ocument {S}ummarization: {A} {P}ilot {S}tudy.
\newblock In \emph{Proceedings of the 11th International Conference on Natural
  Language Generation, Tilburg University, The Netherlands, November 5-8,
  2018}, pages 381--390.

\bibitem[{Zhang et~al.(2018{\natexlab{b}})Zhang, Tan, and Wan}]{Zhang:18}
Jianmin Zhang, Jiwei Tan, and Xiaojun Wan. 2018{\natexlab{b}}.
\newblock Towards a {N}eural {N}etwork {A}pproach to {A}bstractive
  {M}ulti-{D}ocument {S}ummarization.
\newblock \emph{CoRR}, abs/1804.09010.

\bibitem[{Zopf(2018)}]{zopf18mds}
Markus Zopf. 2018.
\newblock Auto-hmds: {A}utomatic {C}onstruction of a {L}arge {H}eterogeneous
  {M}ultilingual {M}ulti-{D}ocument {S}ummarization {C}orpus.
\newblock In \emph{Proceedings of the Eleventh International Conference on
  Language Resources and Evaluation, {LREC} 2018, Miyazaki, Japan, May 7-12,
  2018.}

\end{thebibliography}
\bibliographystyle{acl_natbib}
\end{document}